\documentclass{article}

\usepackage{PRIMEarxiv}

\usepackage[utf8]{inputenc} 
\usepackage[T1]{fontenc}    
\usepackage{hyperref}       
\usepackage{url}            
\usepackage{booktabs}       
\usepackage{amsfonts}       
\usepackage{nicefrac}       
\usepackage{microtype}      
\usepackage{lipsum}
\usepackage{fancyhdr}       
\usepackage{graphicx}       
\graphicspath{{media/}}     

\title{A Roadmap for Embodied and Social Grounding in LLMs
\thanks{Accepted Version of a conference paper presented at Robophilosophy Conference 2024. \textit{\underline{Citation}}: 
\textbf{Incao S., Mazzola C., Belgiovine G., Sciutti A., A Roadmap for Embodied and Social Grounding in LLMs, Robophilosophy Conference, 2024, Aarhus.}} 
}

\author{
  Sara Incao \\ 
  Istituto Italiano di Tecnologia \\
  CONTACT Unit \\
  Genoa (IT)\\
  \textit{Orcid: 0000-0002-5826-520X}
  \thanks{Contact Author: sara.incao@iit.it}
   \And
  Carlo Mazzola \\
  Istituto Italiano di Tecnologia \\
  CONTACT Unit \\
  Genoa (IT)\\
  \textit{Orcid: 0000-0002-9282-9873}
  \And
  Giulia Belgiovine \\
  Istituto Italiano di Tecnologia \\
  CONTACT Unit \\
  Genoa (IT)\\
  \textit{Orcid: 0000-0002-6376-9963}
  \And
  Alessandra Sciutti \\
  Istituto Italiano di Tecnologia \\
  CONTACT Unit \\
  Genoa (IT)\\
  \textit{Orcid: 0000-0002-1056-3398}
}

\begin{document}
\maketitle

\begin{abstract}
The fusion of Large Language Models (LLMs) and robotic systems has led to a transformative paradigm in the robotic field, offering unparalleled capabilities not only in the communication domain but also in skills like multimodal input handling, high-level reasoning, and plan generation. The grounding of LLMs knowledge into the empirical world has been considered a crucial pathway to exploit the efficiency of LLMs in robotics. Nevertheless, connecting LLMs' representations to the external world with multimodal approaches or with robots' bodies is not enough to let them understand the meaning of the language they are manipulating. Taking inspiration from humans, this work draws attention to three necessary elements for an agent to grasp and experience the world. The roadmap for LLMs grounding is envisaged in an active bodily system as the reference point for experiencing the environment, a temporally structured experience for a coherent, self-related interaction with the external world, and social skills to acquire a common-grounded shared experience.
\end{abstract}


\section{Introduction} 

The classic formulation of the symbol grounding problem was proposed by Harnad in 1990 \cite{harnad1990symbol}. He attempted to explain how an artificial agent can autonomously elaborate the semantic aspect of a symbol system, that is, how semantics can be intrinsic to the system, without it being provided from the outside. The symbol grounding problem arose in connection with cognitive science theories about symbolic or computational models of the mind \cite{fodor1983representations, minsky1974framework}. 
According to these models, the mind was conceived as a system manipulating symbols that were representations of objects in the world. In such a view, the use of symbols was dependent upon the application of predefined rules that were totally independent of the physical reality of phenomena and objects represented. Such a purely syntactic system, which is the one digital computers use, was believed to accurately represent the functioning of the mind. It would have been really easy then, to obtain meaning from a syntactic system by finding the right way to connect it to the world, that is, to the physical instantiation of the symbol. Decades after such proposals, the problem of meaning is still an open question, revealing that the idea of a purely formal system connected to the world is not enough. One of the first researchers to address the obstacles of computational models of the mind was John Searle with the “Chinese Room Argument” \cite{searle1980minds}. He argued that a system that provides the right answer to a question only by manipulating properly a symbol system according to known rules does not \textit{understand} the meaning of that answer or of the language the symbols belong to.

Connecting an artificial system to the world is not enough to let it understand the meaning of the symbols it is using. Indeed, if we were looking at an Aramaic word without knowing Aramaic language, that word would not evoke any kind of significance. If we were then shown the object corresponding to that word, that would be helpful only if we were in some sense familiar with that object. If we had never seen it before, it would be impossible for us to understand the \textit{meaning} of the written Aramaic word. In this case, the object would just be another symbol, as the word was before, that is, another referent that brings forth no reference. This shows that the physical reality of objects or, more in general, of the world is insufficient to provide \textit{meaning}. Where does meaning come from then? How is it that human beings are able to “discriminate, manipulate, identify and describe the objects, events, and states of affairs in the world they live in” \cite{harnad1990symbol}? To answer this question, Harnad proposes to analyze the concepts of discrimination and identification closely. \textit{Discrimination} is the ability to judge whether two inputs are the same or different and, in case they are different, what are the elements that distinguish them. \textit{Identification} is instead assigning an arbitrary name to a category of inputs based on them being equivalent or invariant in some aspects. It also involves the ability to discern whether or not a given input is part of a particular category. It could be said that the process of discrimination implies evaluating the degree of similarity of two inputs, and this can be done only by identifying the salient features of those inputs. The issue at hand is why some features of the two inputs are considered more \textit{salient} than others. 
As noted above, the process of meaning-making, given that meaning is not intrinsic to the symbol system used to communicate and it cannot be derived by a simple connection of the symbol system to its physical correspondents, must arise from a different source or type of connection. Meaning, in the first place, emerges as the direct interaction between the agent and the world \cite{sun2000symbol}, that is, as the embodied contact of the subject with the environment in which she lives and acts.

\section{State of the art of grounding with Large Language Models and robotics} 

In the last few years, the increased efficiency and popularity of Large Language Models (LLMs) have reignited the discussion about symbol grounding in artificial systems. The efficiency of LLMs in processing natural language significantly exceeds the performance of previous techniques. The reason for this success is that LLMs are trained on a massive amount of text, which allows for the learning of patterns, associations, and the underlying hierarchical structure of a language. Therefore, LLMs functioning makes it possible to deal with longer-range dependencies in language, to capture hidden contextual and pragmatic information, and to interpret non-literal utterances \cite{mahowald2024dissociating}. Such abilities make them much more advanced than rule-based symbolic systems and previous data-driven machine-learning approaches. Of course, their ability to handle linguistic ambiguity and variability does not mean that LLMs are also able to access the intended meaning and semantic implications of the construction \cite{mahowald2024dissociating}. 

The fusion of LLMs and embodied systems has led to a transformative paradigm in the robotic field offering unparalleled capabilities not only in the communication domain but also in skills like multimodal input handling, high-level reasoning, and plan generation. 
Studies have demonstrated that the effectiveness of these models in text-based tasks has been extended to embodied control \cite{ahn2022i}.
This breakthrough fosters and encourages the introduction of robots in scenarios where co-presence with humans is envisaged. At the same time, the widespread use of such models in interaction contexts highlights the limits of this technology, especially data privacy, ethical considerations, and contextual understanding. 
Different studies have demonstrated, for instance, that LLMs struggle with tasks demanding skills such as physical and social reasoning, which are essential in Human-Robot Interaction (HRI) \cite{ZHANG2023100131}.

The grounding of LLMs knowledge into the empirical world has been considered a crucial pathway to overcome such limitations. 
In the context of LLM-based embodied artificial systems, grounding more precisely refers to the process of aligning the abstract knowledge of foundation models with the tangible, real-world specifics of the agent, ensuring that language-driven decisions correspond meaningfully with actions and interaction contexts  \cite{hu2023robofm}. 

Some researchers have started exploring how robots can enhance their reasoning and manipulation abilities by grounding the physical properties of objects \cite{zhao2023chat}. 
Through direct interaction, a robot could, for instance, understand the concept of a metallic box by processing multimodal cues such as its texture, temperature, weight, and sound produced when tapping on it. In this way, the robot could face uncertainty by proactively acting on the environment and relying on multimodal evidence. 
However, grounding spans diverse applications in the context of robotics.
According to the categorization proposed by \cite{hu2023robofm}, it can refer to the problem of grounding the language or latent concepts (unified latent representations derived from training with diverse input data). 
Moreover, they distinguish between grounding (language or latent concepts) to the environment or to the embodiment. In the former case, the goal is to learn a link between the embedding output and either skills or low-level actions that are strictly related to the environment in which the agent acts.
Grounding to embodiment necessitates instead a specific condition: it must be agnostic to different environments. This is akin to having a universal interface that translates language into actions. For example, grounding latent concepts to embodiment involves directly anchoring foundation models to output the robot's joint torques, circumventing intermediary interfaces such as text \cite{brohan2023rt2}. 

Although numerous strategies have been explored to address the problem of grounding, there are many open challenges in this area \cite{hu2023robofm}. 
First, we need to move from an unimodal notion of grounding, like mapping the word to meaning, to a more holistic grounding of multiple sensory modalities. For example, approaches that rely solely on visual data 
fall short in grasping concepts requiring interaction and proprioceptive feedback.
Moreover, we should consider grounding from an embodiment perspective. The same task may necessitate distinct actions based on the robot’s embodiment; for example, opening a door would require drastically different maneuvers from a humanoid robot compared to a quadruped. Current research on grounding often emphasizes environmental adaptation while affording less consideration to how embodiment shapes interaction strategies.

However, providing LLMs, and even multi-modal LLMs, with robotic bodies endowed with sensors and actuators is far from being sufficient for acquiring, processing, and deploying a grounded knowledge of reality. No meaning can emerge from the association between modalities (e.g., vision and text) or from LLMs maneuvering robots' bodies as puppets. The roadmap for LLMs to grasp reality needs to be anchored to the same core aspects humans rely on. In the following paragraphs, we will analyze the three core steps the development of LLMs-endowed robots should undertake: an active bodily system as the reference point for experiencing the environment, a temporally structured experience for a coherent, self-related interaction with the external world, and social skills to acquire a common-grounded shared experience.

\section{Body and experience}

Robots differ from other artificial systems primarily because they are embodied artificial agents. More precisely, for a cognitive embodied system, the embodiment not only means being implemented in physical form, a feature that belongs to most technological systems. It requires an active role of the body in the cognitive processes and the interaction with the environment \cite{pezzulo2013computational}. In this way, for the artificial cognitive system, the body becomes the means for grounded cognition, where processes linked to perception, action, learning, abstraction, self-monitoring, etc. are all shaped based on the system’s body.


Considering the perspective of embodied cognition \cite{varela2016embodied}, the body is something that cannot be worn and abandoned \cite{MerleauPonty}. Moreover, it is not only an additional property identifying a single person but is essentially and inherently tied to the existence of that agent. \textit{Being a body} indicates that the body is the means, the center, and the basis of experience \cite{MerleauPonty}. It allows the interaction with the world around not merely as an instrument to receive information from the outside and act toward the environment. Human beings are not just passive observers in front of the world, they continuously perceive the environment and act accordingly on it. Even more, our actions shape the way we perceive the world. Our body is engaged in practices we employ every day by means of a combination of perception and action. \cite{varela2016embodied}. 

Learning processes unfold from the very beginning of our life on the background of bodily experience \cite{FalckYtter2006}. 
Specifically, the learning processes of perceptual skills are affected by the development of motor functionality. As regards auditory abilities, the recognition of the minimum audible angle that is necessary to localize a sound improves in the child at around two years of age, in consequence of better control of head movements \cite{Bertenthal1996}. Similarly, considering precision movements such as the gestures of hands and fingers, precision increases with increasing accuracy in discriminating objects’ properties, e.g., size, texture, temperature, weight, shape, etc. \cite{Bushnell1993}. 
Humans can act upon the world and perceive it only as being a body. The body is our anchor to the world, and being a body means not only that we are in space and time. Rather, we inhabit space and time. Therefore, motor experience is our way to get access to the world, that is, to the objects we perceive \cite{MerleauPonty}, and body schema is the non-conscious constant organization responsible for our body's operative performance in the environment \cite{Gallagher1986Bodyschema}.

\textit{Externalist} perspectives in the philosophy of language conceive language as an abstract system where meaning is grasped as an entity that is external to the speaker. In contrast to this view, \textit{internalist} conceptions emphasize the role of embodiment in meaning-making. In particular, these approaches stress the importance of experience conceived as the history of one’s interactions and contact with the world. Such a history is seen as organized in “clouds” of events, objects, situations, and actions whereby meaning connected to words originates \cite{buccino2016grounding}. Becoming a linguistic being, therefore, not only entails the ability to manipulate symbols. These symbols have to be grounded in experience, and experience, for humans, is possible because we are embodied agents that act and modify the world to make sense of it. Furthermore, the process of meaning-making has to be seen not merely as the individual act of one, rather, it is socially constituted in that subjective experience always takes place as part of a shared environment where others live, experience, and act \cite{dipaolo2021enactive}.

\section{Time and experience}

Predictive Processing (PP) accounts conceive perception as an inference process where top-down generative models are used to infer the causes of sensory stimuli. Since the environment is always changing, predictions are constantly updated on the basis of prediction error, that is, the difference between the expected stimulus and the actual one \cite{Clark2013}. Therefore, the act of perception and all experience in general, appears to be a predictive process based on the identification of regularities in past experience that are organized and reused to predict what will happen in the future. Accordingly, PP approaches account for experience in temporal terms \cite{bogota2023can}. Indeed, it is by drawing from the past that it is possible to infer future possibilities. In particular, the past is composed of prior bodily experiences, that is, a situated and specific embodied history of one’s interaction with the world that transforms into the background context for future interactions. Of course, such an interaction is not a passive reception of stimuli. It also involves action toward the environment.
``Active inference'' is a theory that explains the strict relation between perception and action with the same approach of PP. Hence, to reduce the prediction error, its first solution is to find the prediction that better fits reality, whereas the second is to perform actions that make predictions come true \cite{Pezzulo2022}. 
Following this perspective, ``perceptual and motor systems should not be regarded as separate but instead, as a single active inference machine that tries to predict its sensory inputs in all domains'' \cite{Adams2013}. 

Internal models used for predictions are structured frameworks of acquired experience that serve as interpretative scaffolding to internalize new information. They allow for the re-use and generalization of acquired practices to face new situations. Sensory inputs are, therefore, never grasped in isolation: internal models represent the background of every perception. If each sensory perception is intricately linked to its surrounding context, it follows that contexts are shaped through the accumulation and association of various experiences. Contexts serve as the framework through which sensations are interpreted and understood while simultaneously emerging from these very experiences. This relationship can be visualized as a circular dynamic between context and experience \cite{mazzola2022hermeneutical} where the repetition of similar events gradually solidifies into the uniformity of a context, which subsequently becomes customary through experience \cite{mazzola2022human}. Moreover, the presence of multiple perceptions can be integrated and comprehended precisely because they exist within the backdrop of a shared interpretative context. Thus, it becomes evident that perception necessitates both elements: a contextual framework to interpret individual phenomena and the ongoing process of learning from experiences to deepen our understanding of established contexts or to grasp novel ones.

Since the beginning of their life, each person continuously acquires experiences, forming internal models that enable them to predict the sensory-motor-affective consequences of their actions. This process is fundamental in providing the individual with models that shape their interpretation of future events so that novel situations are understood through the lens of existing experience. It is through the formation of experience that humans can grow: new experiences are structured within the framework of previous ones, leading to ``filtering'' the processing of novel events and enabling flexible access and reuse of acquired models to propose novel ways to reach a goal. Past experience allows us to predictively orient our attention toward what matters most and to creatively generalize acquired skills to new situations. For instance, after having experienced firsthand transporting an object, an infant will automatically look at the goal position of other people’s transporting actions rather than at their movement, shifting its attention predictively \cite{falck2006infants}. Furthermore, once adults, the acquisition of a motor skill such as playing an instrument changes the neural regions recruited during perception, leading to a change in auditory perception, attention when listening to music, and even musical imagery \cite{lahav2007action}.

However, also visual experiences can change the understanding of the world, as in the famous ``Dalmatian dog'' illusion: once the dog ``hidden'' in the pattern of black dots forming the image has been seen, it will be forever unveiled for the observer. The experience humans acquire can also be re-used to find creative solutions to unexpected problems. For instance, by reusing the acquired knowledge that both a hand and an elbow can achieve the goal of ``push'', it is possible to realize that a door can be opened with the elbow if the hands are encumbered. Experience formation is only partially determined by the environment surrounding the agent, as it crucially also hinges on the way the individual processes what is happening, in terms of what they are sensitive to, to which aspects they attribute more relevance, to which extent they have already acquired past experiences and to what are their goals and motivations. In particular, the affective dimension plays a central role, as experiences affectively connotated can dramatically influence how events are interpreted, perceived, memorized, and recalled \cite{damasio1994descartes}. 

The term experience related to an artificial agent is usually connected to the possibility of storing information and data collected from the environment and from internal signals such as the position of the limbs or velocity of movements. Besides this operational aspect that proves to be efficient in task-specific circumstances, the design of robots meant to engage in real-time interactions and deal with unexpected situations should take inspiration from human experience in its being temporally consistent, subjectively situated, and interpersonally shaped.

\section{Sociality and shared experience}

Structuring experience based on self and grounding symbols on such self-related, temporally extended, and interconnected information may not be enough for the artificial system to concretely and incrementally ground symbols generated by LLMs. Therefore, taking inspiration from human cognition, another element seems to be required because meaning is always socially and culturally shaped \cite{mead1934mind}. 
For humans, social interaction has been proposed as the default mode of the brain \cite{Hari2015} and the base for the development of high forms of cognitive representations, enabling, for instance, metaphors, dialogic and reflective thinking \cite{tomasello2003key}. Following this idea, social interaction should be considered equally fundamental when it comes to the implementation of artificial cognitive agents. Robots operating in human-populated, and thus social, environments need social skills, and their cognitive processes (perception, action, decision-making, reasoning, etc.) should be shaped and strengthened based on social interactions. The same may be said about grounding. 

Besides others’ attention or perspective on the environment, also their actions toward it may affect one’s perception. As humans, we are able to interpret others’ inner states from how they behave. Others’ actions toward an object may reveal their intentions, beliefs, and affective states about it, as well as some of its properties. For instance, 5 to 7-year-old children are already able to extract relative weight information from observing an actor lifting and carrying a box \cite{Kaiser1984development}. The perception of an actor grasping an object improves the accuracy of judgments about the object’s size \cite{Gori2011direct}. Evidence of others’ actions influencing an observer’s perception of the environment also comes from studies related to the ecological account of affordances. 
Humans do not perceive objects based on their physical properties, e.g., color, texture, shape, and weight, but on their functionality and affordability. Interestingly, this functional perception of the environment may also be directed to others’ affordances (for a review, see \cite{Creem2013relating}), so humans perceive the environment by scaling its features to the action capabilities of other actors \cite{Stoffregen1999perceiving}.

Bottom-up processes are not the only way to come up with action understanding. On the contrary, it has been argued that even in the case of simple sensory stimuli, these models are not sufficient and require a top-down component \cite{Clark2013}. 
The other way for action understanding and behavior interpretation consists of mentalization, a cognitive ability that leads humans to attribute inner states to others. From a developmental point of view, this capability has been hypothesized to originate from more primitive social skills. Early in the first months of life, a process bringing newborns from understanding that other persons are `like me' to the development of social cognitive skills starts \cite{Meltzoff2007likeme}. From the ninth month, infants start considering others as other intentional agents, an understanding they will gradually develop over the years \cite{Tomasello1999}. 
Gradual progress brings infants from joint attention capabilities to perspective-taking \cite{MollMeltzoff2011}. In early childhood, this process reaches a critical step. The children start distinguishing their mental states from those of others, avoiding the mere egocentric attribution of their mental states, hence ascribing mental states to others \cite{MollMeltzoff2011, Tomasello1999}, a capability defined as ``Theory of Mind'' \cite{Premack1978theory}. More specifically, it refers to the ascription of others’ inner states, such as beliefs, intentions, desires, attitudes, and feelings about a target reference.

Within a social context, the grounding appears no longer as an individual activity and reveals its interactive and shared nature. 
In this sense, social interaction appears to be the only means that provides a common, shared meaning to symbols we use to reflect and communicate. Symbol grounding is a matter of social agreement other than concrete grasping of reality. Sharedness is a term with rich and diverse connotations. In this sense, it expresses the agreement and consensus reached by all parties over what is shared that is granted by a bidirectional disclosure of one's intentional relation to the environment. Hence, the common ground between the self and others is found by integrating and comparing the perspectives and experiences of whom one interacts with: the only way to meaningfully communicate and effectively collaborate.

Since we consistently engage in actions aimed at interacting with others or achieving shared goals, it becomes crucial to acknowledge how these dynamics influence our interpretation and comprehension of the world. Thereby, we necessitate a holistic approach to incorporating social and contextual cues into robotic reasoning and decision-making. For instance, when tasked with fetching a specific utensil from a pantry, the robot must consider more than just the physical features of the object. It must also weigh factors tied to (i) the context (e.g., whether the utensil is accessible, the possibility of someone else using the pantry, or its common sharing among household members), (ii) the individuals it is interacting with (e.g., the safety of passing them the utensil based on their age or physical capabilities), and (iii) the communication modality (is the request urgent? Are there other visual cues, e.g. pointing, that can help solve the task quicker?)

Few studies have started investigating how Large Language Models (LLMs) perform in social reasoning and in generalizing on downstream (interaction) tasks they were not explicitly trained for. While they have demonstrated an understanding of specific patterns in human interaction dynamics, it remains uncertain whether they can effectively represent human models \cite{ZHANG2023100131}. 
The primary challenge remains in successfully translating this understanding into real-world behaviors, ultimately impacting effective human-robot interaction. As robots increasingly engage closely with humans, building a shared understanding of social meanings and values is crucial. This shared domain should also extend to communication skills, going beyond the single linguistic modality (as for LLMs) and encompassing various perception channels and expressive cues. Like humans, artificial agents should \textit{interact} to adapt their knowledge to the accumulated experience.

\section{Conclusion}
The considerations made thus far have led to the characterization of symbol grounding as tightly connected with experience in the real world. Exclusively this kind of experience carries the potentiality for some meaning to emerge because meaning implies a direct involvement of the self in the world. Grounding is not simply the connection of a symbol or even a complex system of representations to a physical reality, rather, it is the incarnation of such representations in experience. Our outlined roadmap points to two constitutive elements for experience: body and temporality. LLMs-endowed robots should be designed and developed following this direction, but something is still missing. There is no meaning if not shared. The knowledge of others appears strictly intertwined with the knowledge of the world because we are continuously forced to compare the world as we see it with the way others seem to perceive it. The correspondence between others' reactions and our own to the same stimuli, along with the joint actions we undertake, are the means to move out of a private dimension of meaning toward a shared one.

\bibliographystyle{unsrt}  
\bibliography{robophilosophy}

\begin{thebibliography}{10}

\bibitem{harnad1990symbol}
Stevan Harnad.
\newblock The symbol grounding problem.
\newblock {\em Physica D: Nonlinear Phenomena}, 42(1-3):335--346, 1990.

\bibitem{fodor1983representations}
Jerry~A Fodor.
\newblock {\em Representations: Philosophical essays on the foundations of cognitive science}.
\newblock Mit Press, 1983.

\bibitem{minsky1974framework}
Marvin Minsky.
\newblock A framework for representing knowledge, 1974.

\bibitem{searle1980minds}
John~R Searle.
\newblock Minds, brains, and programs.
\newblock {\em Behavioral and brain sciences}, 3(3):417--424, 1980.

\bibitem{sun2000symbol}
Ron Sun.
\newblock Symbol grounding: a new look at an old idea.
\newblock {\em Philosophical Psychology}, 13(2):149--172, 2000.

\bibitem{mahowald2024dissociating}
Kyle Mahowald, Anna~A Ivanova, Idan~A Blank, Nancy Kanwisher, Joshua~B Tenenbaum, and Evelina Fedorenko.
\newblock Dissociating language and thought in large language models.
\newblock {\em Trends in Cognitive Sciences}, 2024.

\bibitem{ahn2022i}
Michael Ahn, Anthony Brohan, Noah Brown, Yevgen Chebotar, Omar Cortes, Byron David, Chelsea Finn, Chuyuan Fu, Keerthana Gopalakrishnan, Karol Hausman, Alex Herzog, Daniel Ho, Jasmine Hsu, Julian Ibarz, Brian Ichter, Alex Irpan, Eric Jang, Rosario~Jauregui Ruano, Kyle Jeffrey, Sally Jesmonth, Nikhil~J Joshi, Ryan Julian, Dmitry Kalashnikov, Yuheng Kuang, Kuang-Huei Lee, Sergey Levine, Yao Lu, Linda Luu, Carolina Parada, Peter Pastor, Jornell Quiambao, Kanishka Rao, Jarek Rettinghouse, Diego Reyes, Pierre Sermanet, Nicolas Sievers, Clayton Tan, Alexander Toshev, Vincent Vanhoucke, Fei Xia, Ted Xiao, Peng Xu, Sichun Xu, Mengyuan Yan, and Andy Zeng.
\newblock Do as i can, not as i say: Grounding language in robotic affordances, 2022.

\bibitem{ZHANG2023100131}
Ceng Zhang, Junxin Chen, Jiatong Li, Yanhong Peng, and Zebing Mao.
\newblock Large language models for human–robot interaction: A review.
\newblock {\em Biomimetic Intelligence and Robotics}, 3(4):100131, 2023.

\bibitem{hu2023robofm}
Yafei Hu, Quanting Xie, Vidhi Jain, Jonathan Francis, Jay Patrikar, Nikhil Keetha, Seungchan Kim, Yaqi Xie, Tianyi Zhang, Shibo Zhao, Yu-Quan Chong, Chen Wang, Katia Sycara, Matthew Johnson-Roberson, Dhruv Batra, Xiaolong Wang, Sebastian Scherer, Zsolt Kira, Fei Xia, and Yonatan Bisk.
\newblock Toward general-purpose robots via foundation models: A survey and meta-analysis, 2023.

\bibitem{zhao2023chat}
Xufeng Zhao, Mengdi Li, Cornelius Weber, Muhammad~Burhan Hafez, and Stefan Wermter.
\newblock Chat with the environment: Interactive multimodal perception using large language models, 2023.

\bibitem{brohan2023rt2}
Anthony Brohan, Noah Brown, Justice Carbajal, Yevgen Chebotar, Xi~Chen, Krzysztof Choromanski, Tianli Ding, Danny Driess, Avinava Dubey, Chelsea Finn, Pete Florence, Chuyuan Fu, Montse~Gonzalez Arenas, Keerthana Gopalakrishnan, Kehang Han, Karol Hausman, Alexander Herzog, Jasmine Hsu, Brian Ichter, Alex Irpan, Nikhil Joshi, Ryan Julian, Dmitry Kalashnikov, Yuheng Kuang, Isabel Leal, Lisa Lee, Tsang-Wei~Edward Lee, Sergey Levine, Yao Lu, Henryk Michalewski, Igor Mordatch, Karl Pertsch, Kanishka Rao, Krista Reymann, Michael Ryoo, Grecia Salazar, Pannag Sanketi, Pierre Sermanet, Jaspiar Singh, Anikait Singh, Radu Soricut, Huong Tran, Vincent Vanhoucke, Quan Vuong, Ayzaan Wahid, Stefan Welker, Paul Wohlhart, Jialin Wu, Fei Xia, Ted Xiao, Peng Xu, Sichun Xu, Tianhe Yu, and Brianna Zitkovich.
\newblock Rt-2: Vision-language-action models transfer web knowledge to robotic control, 2023.

\bibitem{pezzulo2013computational}
Giovanni Pezzulo, Lawrence Barsalou, Angelo Cangelosi, Martin Fischer, Ken McRae, and Michael Spivey.
\newblock Computational grounded cognition: a new alliance between grounded cognition and computational modeling.
\newblock {\em Frontiers in Psychology}, 3, 2013.

\bibitem{varela2016embodied}
Eleanor Rosch, Lydia Thompson, and Francisco~J Varela.
\newblock {\em The embodied mind: Cognitive science and human experience}.
\newblock MIT press, 2016.

\bibitem{MerleauPonty}
Maurice Merleau-Ponty.
\newblock {\em Phénoménologie de la perception}.
\newblock Librairie Gallimard, 1945.

\bibitem{FalckYtter2006}
Terje Falck-Ytter, Gustaf Gredebäck, and Claes~Von Hofsten.
\newblock Infants predict other people's action goals.
\newblock {\em Nature Neuroscience}, 9:878--879, 7 2006.

\bibitem{Bertenthal1996}
Bennett.I. Bertenthal.
\newblock Origins and early development of perception, action, and representation.
\newblock {\em Annual review of psychology}, 47(1):431--459, 1996.

\bibitem{Bushnell1993}
Emily~W. Bushnell and J.~Paul Boudreau.
\newblock Motor development and the mind: The potential role of motor abilities as a determinant of aspects of perceptual development.
\newblock {\em Child Development}, 64(4):1005--1021, 1993.

\bibitem{Gallagher1986Bodyschema}
Shaun Gallagher.
\newblock Body image and body schema: A conceptual clarification.
\newblock {\em The Journal of mind and behavior}, pages 541--554, 1986.

\bibitem{buccino2016grounding}
Giovanni Buccino, Ivan Colag{\`e}, Nicola Gobbi, and Giorgio Bonaccorso.
\newblock Grounding meaning in experience: A broad perspective on embodied language.
\newblock {\em Neuroscience \& Biobehavioral Reviews}, 69:69--78, 2016.

\bibitem{dipaolo2021enactive}
Ezequiel~A Di~Paolo.
\newblock Enactive becoming.
\newblock {\em Phenomenology and the Cognitive Sciences}, 20(5):783--809, 2021.

\bibitem{Clark2013}
Andy Clark.
\newblock Whatever next? predictive brains, situated agents, and the future of cognitive science.
\newblock {\em Behavioral and Brain Sciences}, 36:181--204, 2013.

\bibitem{bogota2023can}
Juan~Diego Bogot{\'a}.
\newblock Can the predictive mind represent time? a critical evaluation of predictive processing attempts to address husserlian time-consciousness.
\newblock {\em Phenomenology and the Cognitive Sciences}, pages 1--21, 2023.

\bibitem{Pezzulo2022}
Giovanni Pezzulo, Thomas Parr, and Karl Friston.
\newblock The evolution of brain architectures for predictive coding and active inference.
\newblock {\em Philosophical Transactions of the Royal Society B: Biological Sciences}, 377(1844):20200531, 2022.

\bibitem{Adams2013}
Rick~A Adams, Stewart Shipp, and Karl~J Friston.
\newblock Predictions not commands: active inference in the motor system.
\newblock {\em Brain Structure and Function}, 218(3):611--643, 2013.

\bibitem{mazzola2022hermeneutical}
Carlo Mazzola, Sara Incao, Massimo Marassi, Francesco Rea, Alessandra Sciutti, et~al.
\newblock A hermeneutical approach to provide robots with socially adaptive perception.
\newblock In Raul Hakli, Pekka M{\"a}kel{\"a}, and Johanna Seibt, editors, {\em Social Robots in Social Institutions. Proceedings of Robophilosophy 2022}, volume 366, pages 344--352. IOS Press, 2022.

\bibitem{mazzola2022human}
Carlo Mazzola, Sara Incao, Francesco Rea, Alessandra Sciutti, Massimo Marassi, et~al.
\newblock Human experience and robotic experience. a reciprocal exchange of perspectives.
\newblock In {\em Humane Robotics. A Multidisciplinary Approach towards the Development of Humane-Centered Technologies}, pages 51--68. Vita e Pensiero, 2022.

\bibitem{falck2006infants}
Terje Falck-Ytter, Gustaf Gredeb{\"a}ck, and Claes Von~Hofsten.
\newblock Infants predict other people's action goals.
\newblock {\em Nature neuroscience}, 9(7):878--879, 2006.

\bibitem{lahav2007action}
Amir Lahav, Elliot Saltzman, and Gottfried Schlaug.
\newblock Action representation of sound: audiomotor recognition network while listening to newly acquired actions.
\newblock {\em Journal of Neuroscience}, 27(2):308--314, 2007.

\bibitem{damasio1994descartes}
Antonio Damasio.
\newblock Descartes’ error: Emotion, rationality and the human brain.
\newblock {\em New York: Putnam}, 352, 1994.

\bibitem{mead1934mind}
George~Herbert Mead et~al.
\newblock {\em Mind, self, and society}, volume 111.
\newblock University of Chicago press Chicago, 1934.

\bibitem{Hari2015}
Riitta Hari, Linda Henriksson, Sanna Malinen, and Lauri Parkkonen.
\newblock Centrality of social interaction in human brain function.
\newblock {\em Neuron}, 88(1):181--193, 2015.

\bibitem{tomasello2003key}
Michael Tomasello.
\newblock The key is social cognition.
\newblock In Dedre Gentner and Susan Goldin-Meadow, editors, {\em Language in mind: Advances in the study of language and thought}, pages 47--57. MIT Press, 2003.

\bibitem{Kaiser1984development}
Mary~Kister Kaiser and Dennis~R Proffitt.
\newblock The development of sensitivity to causally relevant dynamic information.
\newblock {\em Child Development}, pages 1614--1624, 1984.

\bibitem{Gori2011direct}
Monica Gori, Alessandra Sciutti, David Burr, and Giulio Sandini.
\newblock Direct and indirect haptic calibration of visual size judgments.
\newblock {\em PLoS One}, 6(10):e25599, 2011.

\bibitem{Creem2013relating}
Sarah~H. Creem-Regehr, Kyle~T Gagnon, Michael~N Geuss, and Jeanine~K Stefanucci.
\newblock Relating spatial perspective taking to the perception of other's affordances: Providing a foundation for predicting the future behavior of others.
\newblock {\em Frontiers in Human Neuroscience}, 7:596, 2013.

\bibitem{Stoffregen1999perceiving}
Thomas~A. Stoffregen, Kathleen~M Gorday, Yang-Yi Sheng, and Steven~B Flynn.
\newblock Perceiving affordances for another person's actions.
\newblock {\em Journal of Experimental Psychology: Human Perception and Performance}, 25(1):120, 1999.

\bibitem{Meltzoff2007likeme}
Andrew~N. Meltzoff.
\newblock ‘like me’: a foundation for social cognition.
\newblock {\em Developmental Science}, 10(1):126--134, 2007.

\bibitem{Tomasello1999}
Michael Tomasello.
\newblock {\em The Cultural Origins of Human Cognition}.
\newblock Harvard University Press, 1999.

\bibitem{MollMeltzoff2011}
Henrike Moll and Andrew~N. Meltzoff.
\newblock Joint attention as the fundamental basis of understanding perspectives.
\newblock In Axel Seeman, editor, {\em Joint attention: New developments in psychology, philosophy of mind, and social neuroscience}, chapter~15, pages 393--413. MIT Press, Oxford, 2011.

\bibitem{Premack1978theory}
David Premack and Guy Woodruff.
\newblock Does the chimpanzee have a theory of mind?
\newblock {\em Behavioral and brain sciences}, 1(4):515--526, 1978.

\end{thebibliography}

\end{document}